%% file: main.tex
\documentclass[10pt,twocolumn,letterpaper]{article}

\usepackage{cvpr}
\usepackage{times}
\usepackage{epsfig}
\usepackage{graphicx}
\usepackage{amsmath}
\usepackage{amssymb}
\usepackage{multirow}
\usepackage{tabularx, makecell}

\usepackage[breaklinks=true,bookmarks=false]{hyperref}

\cvprfinalcopy 

\def\cvprPaperID{6366} 

\ifcvprfinal\fi

\begin{document}

\title{Machine Vision Guided 3D Medical Image Compression for Efficient Transmission and Accurate Segmentation in the Clouds}

\author{\small
Zihao Liu$^1$, 
Xiaowei Xu$^2$, 
Tao Liu$^1$,
Qi Liu$^1$, 
Yanzhi Wang$^3$,
Yiyu Shi$^2$, 
Wujie Wen$^1$,
Meiping Huang$^4$, 
Haiyun Yuan$^4$,
Jian Zhuang$^4$ \\
\small $^1$ Flordia International University,
$^2$ University of Notre Dame,
$^3$ Northeastern University,
$^4$ Guangdong General Hospital\\
\{\tt\small zliu021,tliu023,qliu020,wwen\}@fiu.edu,
\{\tt\small xxu8,yshi4\}@nd.edu,
 \{\tt\small yanz.wang\}@northeastern.edu
}


\maketitle

\begin{abstract}

Cloud based medical image analysis has become popular recently due to the high computation complexities of various deep neural network (DNN) based frameworks and the increasingly large volume of medical images that need to be processed. It has been demonstrated that for medical images the transmission from local to clouds is much more expensive than the computation in the clouds itself. Towards this, 3D image compression techniques have been widely applied to reduce the data traffic. However, most of the existing image compression techniques are developed around human vision, i.e., they are designed to minimize distortions that can be perceived by human eyes. In this paper we will use deep learning based medical image segmentation as a vehicle and demonstrate that interestingly, machine and human view the compression quality differently. Medical images compressed with good quality w.r.t. human vision may result in inferior segmentation accuracy. We then design a machine vision oriented 3D image compression framework tailored for segmentation using DNNs.
Our method automatically extracts and retains image features that are most important to the segmentation. Comprehensive experiments on widely adopted segmentation frameworks with HVSMR 2016 challenge dataset show that our method can achieve significantly higher segmentation accuracy at the same compression rate, or much better compression rate under the same segmentation accuracy, when compared with the existing JPEG 2000 method. 
To the best of the authors' knowledge, this is the first machine vision guided medical image compression framework for segmentation in the clouds. 


\end{abstract}

\input{1.intro.tex}

\input{2.back.tex}
\input{3.design.tex}

\input{5.evaluation.tex}

\input{6.conclusion.tex}

{\small
\bibliographystyle{ieee}
\bibliography{egbib}
}

\end{document}

%% file: 1.intro.tex
\begin{figure*}[!h]
\centering
 \includegraphics[width=1.9\columnwidth]{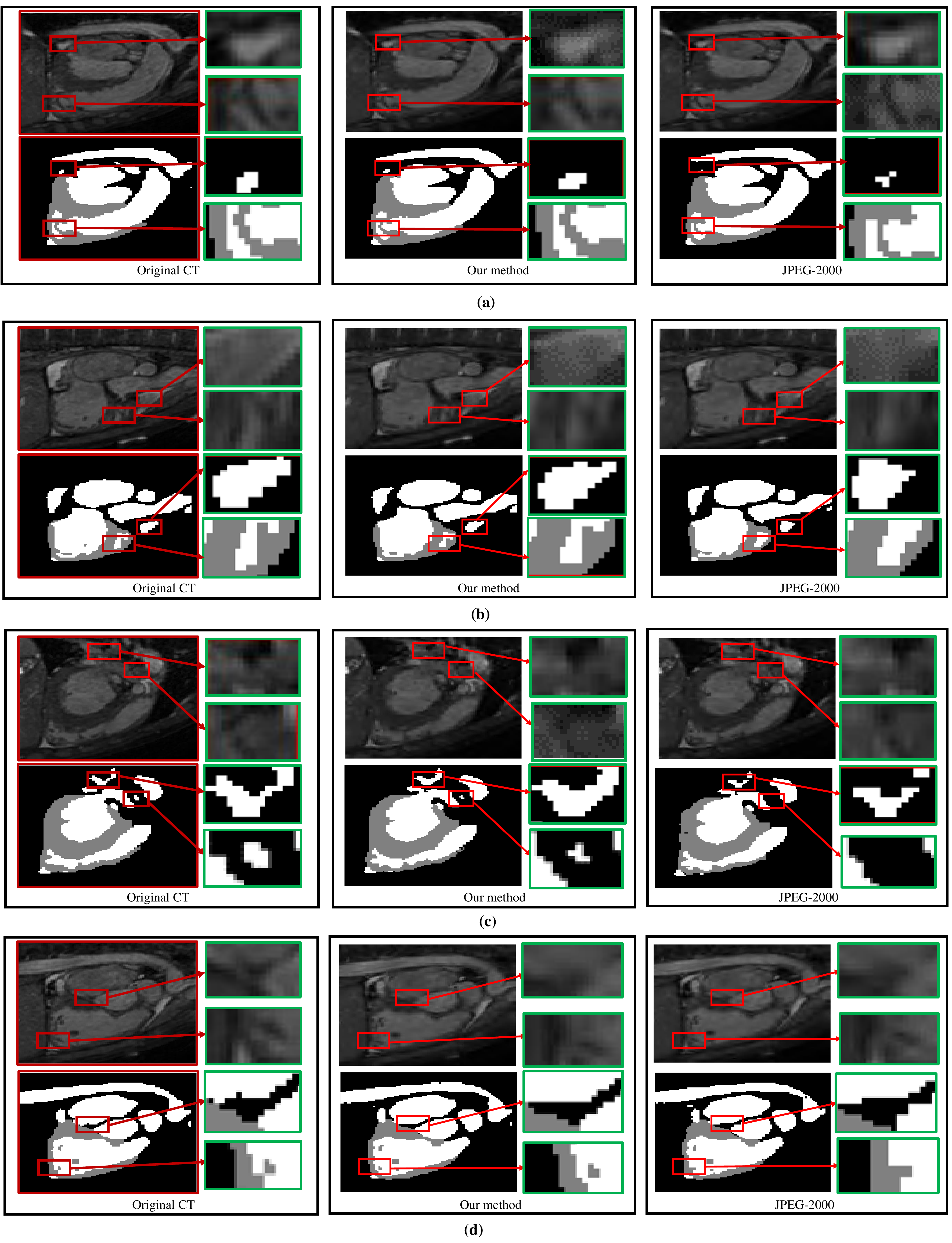}
 \caption{Segmentation details of four slices in a CT image in HVSRMR 2016 Challenge dataset \cite{pace2015interactive}, compressed using our method and JPEG-2000, and segmented by DenseVoxNet~\cite{yu2017automatic}. 
 Many details are missing in the segmentation results from JPEG-2000 compressed images but not in our method. Quantitative comparisons can be found in Section~\ref{sec:exp}.}
 \label{motivationFig}
\end{figure*}
\section{Introduction}
\label{int}

Deep learning has significantly pushed forward the frontier of automatic medical image analysis \cite{cciccek20163d}\cite{ronneberger2015u}\cite{xu2018quantization}\cite{chen2016deep}\cite{cciccek20163d}\cite{milletari2016v}\cite{chen2017voxresnet}\cite{dou20163d}\cite{tianchen2019}. On the other hand, most deep learning based frameworks have high computation complexities \cite{xu2018scaling}\cite{xu2018resource}\cite{xu2017edge}\cite{xu2017efficient}\cite{xu2018efficient}\cite{jiang2019accuracy}\cite{jiang2019xfer}. For example, the number of operations needed by the network by \cite{chen2016combining} to segment a 3D Computed Tomography (CT) volume would be around 2.2 Tera ($10^{12}$) , which needs days to be processed on a general desktop computer. In addition,  with the advances in medical imaging technologies, the related data has been increasing exponentially for decades \cite{dinov2016volume}. Ponemon Institute survey found that 30\% of the world’s data storage resides in the healthcare industry by 2012 \cite{Jim14}. For both reasons, clouds have become a popular platform for efficient deep learning based medical image analysis \cite{marwan2017using}\cite{zhao2014cloud}\cite{xu2018accelerating}\cite{xu2018mda}\cite{zhao2013cloud}.  

Utilizing clouds, however, requires medical images to be transmitted from local to servers. Compared with computation time needed to process these images in the clouds, the transmission time is usually higher.
For example, the latency to transmit a 3D CT image of size 300MB is about 13 seconds via fixed broadband internet (estimated with 2017 U.S. average fixed broadband upload speed of 22.79 Mbps \cite{networkSpeed}). On the other hand, it takes no more than 100 milliseconds for 3D-DSN \cite{dou20163d} to segment an image through  
a high-performance cluster of 10 GPUs in cloud \cite{ma2017evolution}\cite{coomans2015xg}\cite{kang2017neurosurgeon}. For slower internet speed, this gap is even bigger.

To tackle this issue, image compression is typically used to prune unimportant information before sending the image to clouds,
thus reducing data traffic. The compression time is usually negligible (e.g., 24 milliseconds to compress a 300MB 3D CT image to 30MB using a moderate GPU \cite{matela2009gpu}). 
There exist many general image compression standards such as JPEG-2000 \cite{boliek2002jpeg}\cite{boliek2002jpeg3d}, JPEG \cite{wallace1992jpeg}, and MPEG2 \cite{itu1995generic}.
Most of these standards use frequency transformation 
to filter out information that leads to little visual distortion. 
In addition to the existing 3D image compression standards, alternative compression methods have been proposed in the literature, most of which
modify the existing standards to improve their performance \cite{bruylants2015wavelet}\cite{sanchez20103}\cite{sanchez2009symmetry}\cite{xu2016diagnostically}. There are also a few methods for lossless compression of 3D medical images \cite{santos2015contributions}\cite{lucas2017lossless}. 

Almost all the existing compression techniques are optimized for the Human-Visual System (HVS), or image quality perceived by humans. However, when we compress images for transmission to the clouds, their quality will not be judged by human vision, but rather by the performance of the neural networks that process them in the clouds. As such, an interesting question that naturally arises is: are the existing compression techniques still optimal for these neural networks, i.e., in terms of ``machine visions''? In this paper, we will use 3D medical image segmentation as a vehicle to study this question. 

Medical image segmentation extracts different tissues, organs, pathologies, and biological structures to support medical diagnosis, surgical planning and treatments. We adopt JPEG-2000 to compress the HVSMR 2016 Challenge dataset~\cite{pace2015interactive}, and two state-of-the-art neural networks--DenseVoxNet~\cite{yu2017automatic} and 3D-DSN~\cite{dou20163d} for medical image segmentation. 
The results for four randomly selected slices are shown in Fig.~\ref{motivationFig}. From the figure we can see that quite significant differences exist between the segmentation results from the original image and the one compressed by JPEG-2000, though visually little distortions exist between the two. 

The results may seem surprising at first glance, but it is also fully justifiable. The boundaries in medical images mainly contribute to the high frequency details, which cannot be perceived by human eyes. As such, existing compression techniques will ignore them while still attaining excellent compression quality. Yet these details are critical features that neural networks need to extract to accurately segment an image. Similarly, many low frequency features in a medical image such as brightness of a region are important for human vision guided compression, but not at all for segmentation. In other words, human vision and machine vision are completely different with regard to the segmentation task. 


In this paper, we propose a machine vision guided 3D image compression framework tailored for deep learning based medical image segmentation in the clouds.
Different from most existing compression methods that take human visual distortion as guide, our method extracts and retains features that are most important to segmentation, so that the segmentation quality can be maintained.
We conducted comprehensive experiments on two widely adopted segmentation frameworks (DenseVoxNet~\cite{yu2017automatic} and 3D-DSN~\cite{dou20163d} using the HVSMR 2016 Challenge dataset ~\cite{pace2015interactive}. Examples on the qualitative effect of our method on the final segmentation results can be viewed in Fig.~\ref{motivationFig}.

The main contributions of our work are as follows:
\begin{itemize}
    \item We discovered that for medical image segmentation in the clouds, traditional compression methods guided by human vision will result in inferior accuracy, and a new method guided by machine vision is warranted.
    \item We proposed a method that can automatically extract important frequencies for neural network based image segmentation,
    and map them to quantization steps for better compression. 
    \item Experimental results 
    show our method outperforms JPEG-2000 in two aspects: for a same compression rate, our method achieves significantly improved segmentation accuracy; for a same level of segmentation accuracy, it offers much higher compression rate ($3\times$). These advantages demonstrate great potentials for its application in today's deep neural network assisted medical image segmentation.  
    
    
    

\end{itemize}

%% file: 2.back.tex
\section{Related Work}
\label{pre}
\subsection{3D Medical Image Compression}

There are many general image compression standards such as JPEG-2000 \cite{boliek2002jpeg}\cite{boliek2002jpeg3d}, JPEG \cite{wallace1992jpeg}.
Some video coding standards such as H.264/AVC \cite{telecom2003advanced},and MPEG2 \cite{itu1995generic} can also be adopted for 3D image segmentation.
Most of these standards use transforms such as Discrete Cosine Transform (DCT) and Discrete Wavelet Transform (DWT) for compression while preserving important visual information for humans. 

In addition to the existing 3D medical image compression standard, alternative compression methods have been proposed in the literature.
Most of the methods modified the existing standards to improve its performance.
Bruylants \etal \cite{bruylants2015wavelet} adopted volumetric wavelets and entropy-coding to improve the compression performance.
Sanchez \etal \cite{sanchez20103} employed a 3-D integer wavelet transform to perform column of interest coding.
Sanchez \etal \cite{sanchez2009symmetry} reduced the energy of the sub-bands by exploiting the anatomical symmetries typically present in structural medical images. 
Zhongwei \etal \cite{xu2016diagnostically} improved the compression performance by removing unimportant image regions not required for medical diagnosis.
There are a few methods for lossless compression of 3D medical images.
Santos \etal \cite{santos2015contributions} processed each frame sequentially and using 3D predictors based on the previously encoded frames.
Lucas \etal \cite{lucas2017lossless} further adopted 3D block classification to process the data at the volume level.

Almost all the above methods still adopt the same objective as that used by JPEG-2000, i.e., to minimize human perceived distortions. As shown in the example in Fig.~\ref{motivationFig}, when it comes to the deep learning based segmentation, such a strategy may lead to poor accuracy. 

\begin{figure}[!h]
\centering
\vspace{-10pt}
 \includegraphics[width=0.49\textwidth]{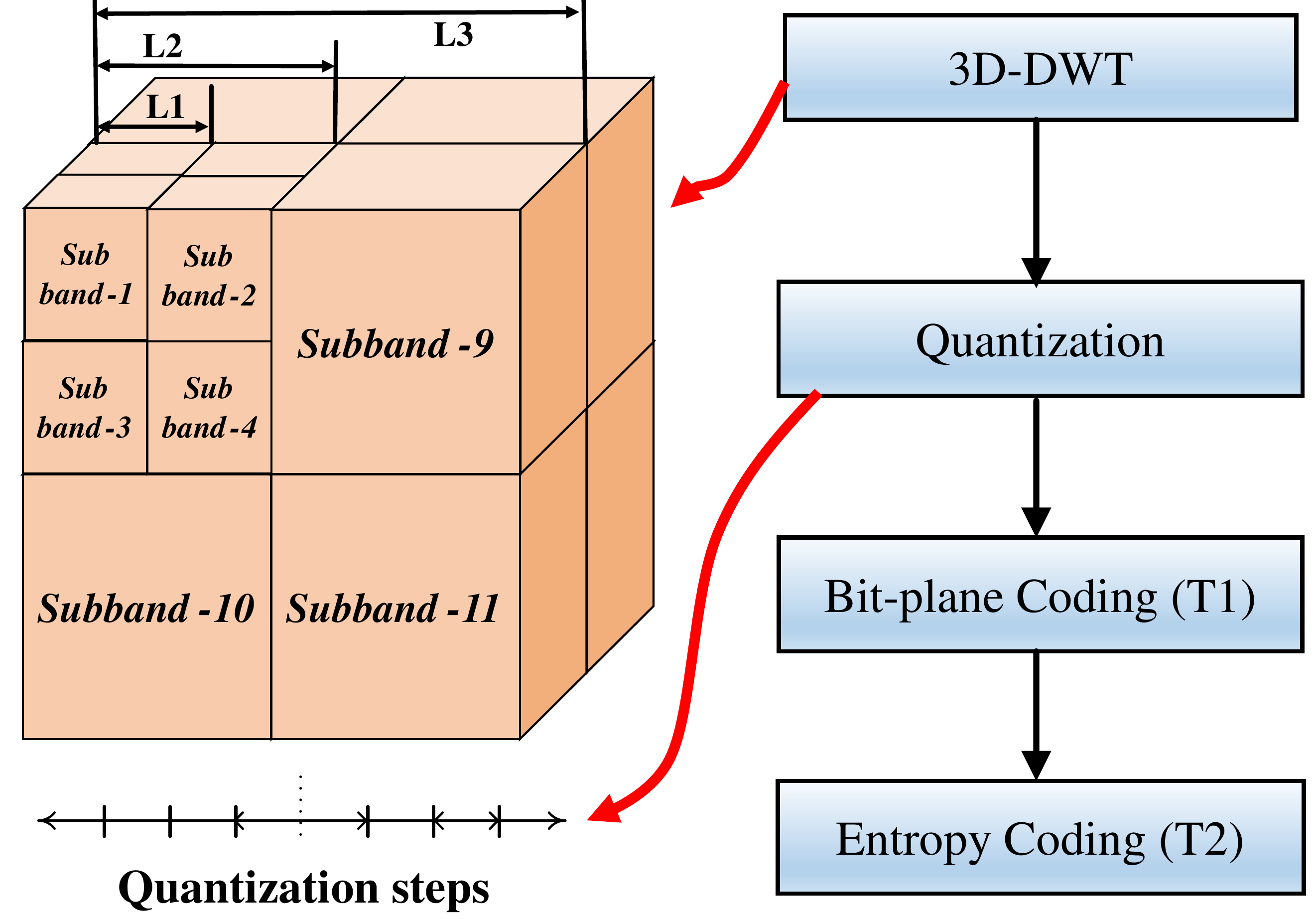}
 \caption{Flow of JPEG-2000 compression method. }
 \label{j2k}
 \vspace{-10pt}
\end{figure}

\subsection{JPEG-2000 3D image Compression}
Our method is also based on JPEG-2000 but modifies its human vision guided objective to one that is guided by the segmentation network. Here we briefly review the details of JPEG-2000 so that later we can explain our work better. 
Fig.~\ref{j2k} shows the major steps in JPEG-2000 compression: 
First, the 3D discrete wavelet transform (DWT) is applied to an image to decompose it into a multiple-resolution representation in frequency domain~\cite{shensa1992discrete}\cite{antonini1992image}\cite{penna2006progressive}. For example, a 3-D wavelet decomposition leads to three resolution levels (L1, L2, L3). Each resolution level (except L1) is composed of eight subbands: subband 1 to subband 8.
The eight lower resolution levels are always generated by progressively applying the 3D DWT process to the upper-left-front block (e.g., subband 1) from the previous resolution level. Then a non-uniform quantization process is applied to each subband based on the number of low pass filters in the subband:

\begin{equation}
x'=\lfloor\frac{x}{QS}\rfloor
\end{equation}
where $x$ is the original coefficient after 3D DWT, $QS$ is the quantization step of a subband and $x'$ is the coefficient after quantization. 

The rule is that the more low pass filters a subband has the smaller quantization step are applied to the corresponding subband. This is because Human Visual System (HVS) is more sensitive to low pass frequency information, thus less quantization errors in low pass subband. 
Bit-plane coding and entropy coding mainly perform coding and please interested readers are referred to the related literature \cite{taubman2000high}\cite{schwartz1966bit}\cite{skodras2001jpeg}\cite{taubman2000high} for more details. 

%% file: 3.design.tex
\section{Machine Vision oriented 3D Image Compression}

\begin{figure*}[t]
\centering
 \includegraphics[width=1\textwidth]{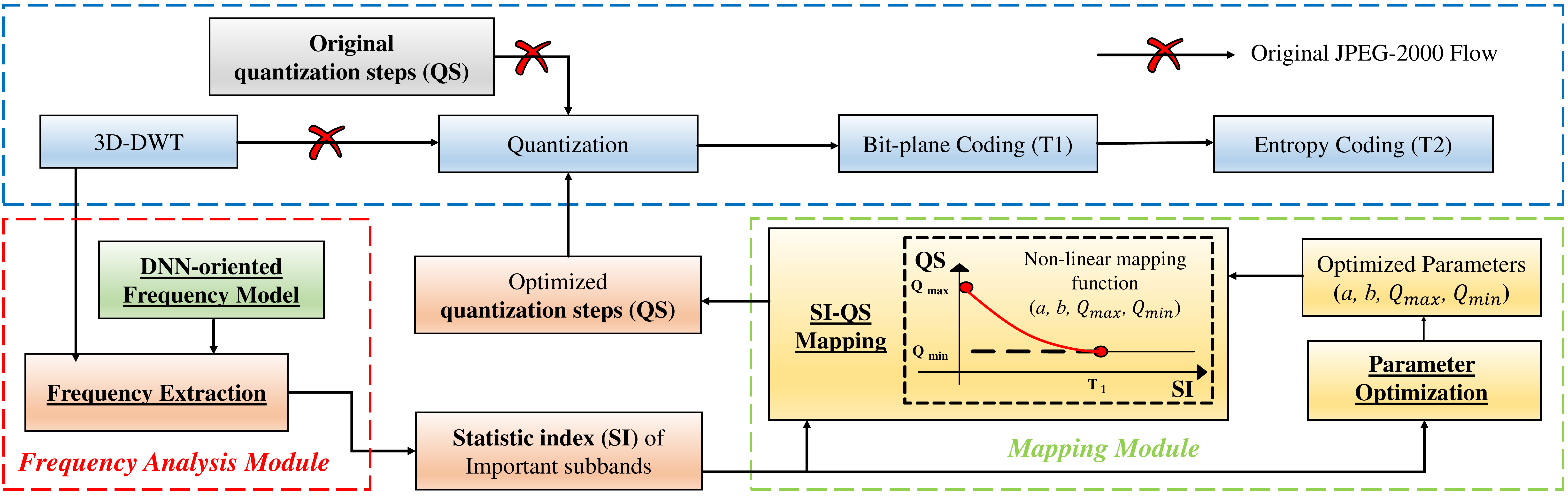}
 \caption{Overview of the proposed DNN-oriented 3D image compression framework.}
 \label{flow}
\end{figure*}

In this section, the details of the proposed machine vision oriented 3D image compression framework for segmentation in the clouds is presented. As shown in Fig. \ref{flow}, the framework contains two modules: frequency analysis module and mapping module.
Compared with original JPEG-2000 compression method, the added two modules can obtain optimized quantization steps (QSs) for better segmentation accuracy.
The frequency analysis module extracts frequencies important to segmentation with high statistic indexes (SI) using a machine vision guided frequency model.
The mapping module maps these SIs to optimized QSs which are further provided to the quantization module in JPEG-2000 flow for the rest of the processing. 
Particularly parameter optimization is also performed to find the optimal parameters in the mapping module.



\subsection{Frequency Analysis Module}

\subsubsection{Machine Vision Guided Frequency Model}
In this section we build a frequency model that identifies information most useful for segmentation.
Assume $x_i$ is a single voxel of a raw 3D image \textbf{X}. $x_i$ can be represented by 3D-DWT at one resolution level in JPEG-2000 compression as: 

\begin{equation}
x_i=\sum_{n=0} ^{i=N-1} c_{i}^n\cdot b_i^n 
\end{equation}
where $c_i^n$ and $b_i^n$ are the 3D-DWT coefficient at matching 3D coordinate $i$ and corresponding basis function at $N$ different subbands, respectively. 

For human visual system, the quantization step (QS) for each subband in JPEG-2000 is positively correlated with the number of high pass filters in a subband.
For example, the QS of 
subband 4 is larger than that of 
subband 2 at the same resolution level. Then larger QS in high frequency subband will increase the distortion of coefficients in this subband. Consequently, it will either directly zero out the associated 3D-DWT coefficient $c_i^n$ or increase the chance to truncate them at rate-distortion optimization process. This is because HVS is less sensitive to high frequency subband, so a high compression rate can be achieved by discarding the high frequency information. 

In order to obtain the important frequency for DNN based segmentation, we calculate the gradient of the DNN loss function F with respect to a basis function $b_i^n$ as:

\begin{equation}
{ \frac{\partial F}{\partial b_i^n}=\frac{\partial F}{\partial x_i}\times\frac{\partial x_i}{\partial b_i^n}=\frac{\partial F}{\partial x_i}\times c_{i}^n. }
\label{basis}
\end{equation}

Equation ~(\ref{basis}) indicates that the importance of information at different subbands of a single voxel to DNN is determined by its associated 3D-DWT coefficients ($c_i^n$) at all subbands. This is quite different from HVS which distorts $c_i^n$ in high frequency subbands (i.e. quantization or truncation). Large $c_i^n$ in high frequency subband will be heavily distorted in JPEG-2000. However, it may carry important information for DNN segmentation, causing accuracy degradation. 

\subsubsection{Frequency Extraction}



\begin{figure}[!b]
\centering
\vspace{-10pt}
 \includegraphics[width=0.95\columnwidth]{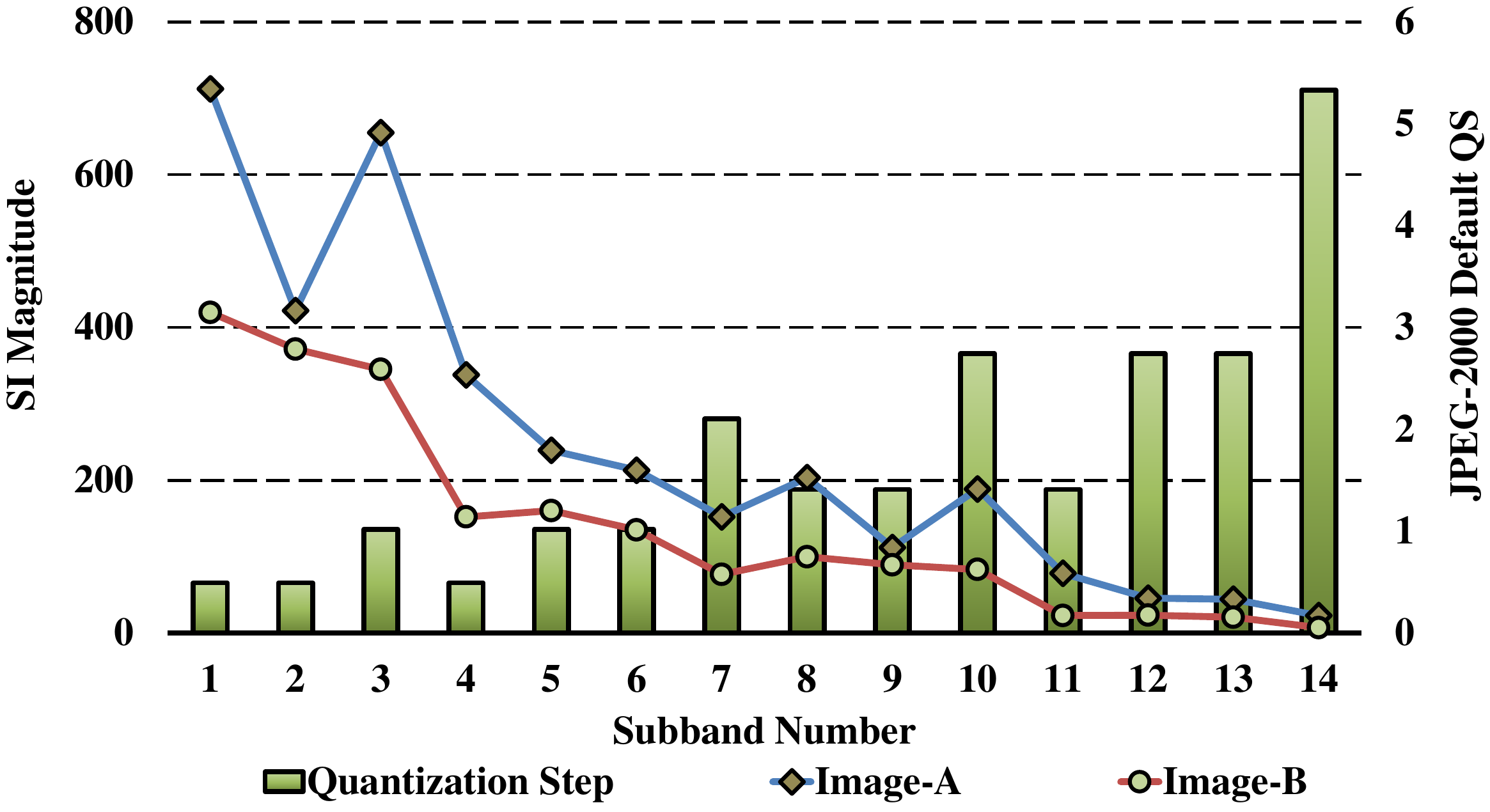}
 \caption{Diverse frequency domain of medical images.}
 \label{fre}
\end{figure}

In this section, we extract important frequencies based on the above frequency model.
Previous studies~\cite{stamm2011anti}\cite{li1998text} have demonstrated that the distribution of un-quantized 3D-DWT coefficients in a subband indicates the energy in this subband. Moreover, the distribution of each subband has been proven that they approximately obey a Laplace distribution with zero mean and different standard deviations ($\delta_n$). The larger $\delta_n$ a subband has (i.e. more energy in this subband), the more contribution this subband will provide to DNN results. Therefore, $\delta_n$ of each subband after 3D-DWT are selected as the SI to represent the importance to DNN. Based on this we propose to conduct the frequency analysis as follows: 
the number of subbands will be first calculated based on the number of resolution levels at three different dimensions provided by users. After that coefficients that belong to the same subband will be grouped up and reshaped to one dimension. Then the distributions of reshaped coefficients at each subband will be characterized. Finally, the statistical information of each subband, i.e. the standard deviation or SI, will be calculated based on its histogram. The results from this frequency information projection procedure can clearly indicate the importance of each subband to DNN by its SI. 
With the above discussion, we further analyzed SIs and QSs in JPEG-2000 to show that JPEG-2000 is not optimized for DNNs.
We randomly selected two images from HVSMR 2016 dataset labeled as A and B, and then applied our frequency extraction method on them after 3-3-3 3D-DWT.
As shown in Fig.~\ref{fre}, 
some important subbands have large QS which is undesired. For example, subband 2 is less important than subband 3 for image A since $\delta_2<\delta_3$, however, its QS is much smaller than that of subband 3. 
The same problem exists for subband 3 and subband 4 with image B. 
Thus, although lower frequency information is always more important than that of higher frequency in JPEG-2000, it is not the case for segmentation accuracy.
\subsection{Mapping Module}

\subsubsection{SI-QS Mapping}
With SIs at each subband, our next step is to find a suitable mapping between SI and QS by well leveraging the intrinsic error resilience characteristic of DNN computation. As a result, the segmentation accuracy loss due to increasing compression rate, can be minimized by largely quantilizing the frequency subbands that are less significant to DNN.






In order to precisely model the mapping, we attempt to find a QS curve aligning with most of the SIs. 
With extensive experiments (we add these experiments in the supplemental material), we observe that the QS-SI points obey a reciprocal function ($y=1/x$).
Thus, we propose a non-linear mapping (NLM) method to implement nonuniform quantization steps at different subbands: 

\begin{equation}
\label{nlm}
Q_n=\frac{a}{(\delta_n+b)}, \quad s.t. \quad Q_{min}\leq Q_n \leq Q_{max}
\end{equation}
where $Q_n$ is the quantization step at subband $n$, $Q_{min}$ and $Q_{max}$ are the smallest and largest QS, and $a$ and $b$ are the fitting parameters.

\begin{table*}[t]
\centering
 \caption{Segmentation results of our methods and JPEG-2000 using DenseVoxNet and HVSMR2016 dataset. The compression rate is set to 30 for both techniques. The images compressed by ours can be segmented with almost the same accuracy as, or sometimes even better than the original ones, much better than those compressed by JPEG-2000. 
 The segmentation performance of NLM is very close to or even better than that with the original images while is much better than JPEG-2000. }
 \label{dens}
\begin{tabularx}{\textwidth}{XXXXX}
\Xhline{3\arrayrulewidth}
\multicolumn{2}{c}{}                  & \textbf{Original}               & \textbf{Ours}    & \textbf{JPEG-2000}                 \\
\Xhline{3\arrayrulewidth}
\multirow{3}{*}{\textbf{Myocardium}} & \textbf{Dice}  & 0.838$\pm$0.0334   & \textbf{0.834$\pm$0.0386}  & 0.816$\pm$0.042 \\
                            & \textbf{Hausdorff} & 30.879$\pm$7.592  & \textbf{31$\pm$7.940}     
                            & 33.513$\pm$7.566 \\ 
                            & \textbf{ASD} & 0.673$\pm$0.67  & \textbf{0.652$\pm$0.671}      
                            & 0.722$\pm$0.746  \\ 
\Xhline{2\arrayrulewidth}  
\multirow{3}{*}{\textbf{Blood Pool}} & \textbf{Dice}  & 0.915$\pm$0.025 & \textbf{0.914$\pm$0.024}  
& 0.912$\pm$0.025 \\
                            & \textbf{Hausdorff} & 41.034$\pm$9.326   & \textbf{40.93$\pm$9.52}  
                            & 41.031$\pm$9.648 \\
                            & \textbf{ASD} & 0.601$\pm$0.455   & \textbf{0.556$\pm$0.432} 
                            & 0.582$\pm$0.453         \\ 
\Xhline{3\arrayrulewidth}
\multicolumn{2}{c}{\textbf{Compression Rate}}  & 1      & \textbf{$\sim$30x} 
& \textbf{$\sim$30x}      \\
\multicolumn{2}{c}{\textbf{PSNR (dB)}}  & $\infty$ & $\sim$35 
& \textbf{$\sim$36}  \\
\Xhline{3\arrayrulewidth}
\end{tabularx}
\end{table*}

\subsubsection{Parameter Optimization}\label{paraOpt}
With the proposed mapping function, parameter optimization is performed to obtain the optimal $a$, $b$, $Q_{max}$ and $Q_{min}$ in Equation (\ref{nlm}). 
For $a$ and $b$, we found that rational functions can fit the relationship between the standard deviation of each subband of an image and the quantization step very well. For $Q_{max}$ and $Q_{min}$, 
we examine two corner cases, i.e. upper/lower corner to explore the quantization error tolerance for the most insignificant/significant subband.
Then all the parameters 
in non-linear mapping method can be calculated by substituting pairs ($Q_{min}$, $\delta_{max}$) and ($Q_{max}$, $\delta_{min}$) into Equation~(\ref{nlm}). 


\textbf{Lower Corner Case:}
we assign the same QS to all the subbands to explore $Q_{min}$.
As long as the error induced by QS in the subband with $\delta_{max}$ (the most significant subband to DNN) does not impact the segmentation accuracy, this will also hold true for all the other subbands.

\textbf{Upper Corner Case:}  
To find $Q_{max}$, we only vary QS at the subband with $\delta_{min}$, while fixing that of all the other subbands as the same QS--$Q_{min}$.
If 
the subband with $\delta_{min}$ (the least significant subband to DNN) cannot tolerate the error incurred by a $Q_{max}$, the other subbands cannot either.

%% file: 5.evaluation.tex
\begin{table*}[t]
\centering
 \caption{Segmentation results of our methods and JPEG-2000 using 3D-DSN and HVSMR 2016 dataset. The compression rate is set to 30 for both techniques. The images compressed by ours can be segmented with almost the same accuracy as the original ones, and significantly better than those compressed by JPEG-2000. }
 \label{dsn}
\begin{tabularx}{\textwidth}{XXXXXXX}
\Xhline{3\arrayrulewidth}
\multicolumn{2}{c}{}                  & \textbf{Original}                                   & \textbf{Ours}                        & \textbf{JPEG-2000}             \\
\Xhline{3\arrayrulewidth}
\multirow{3}{*}{\textbf{Myocardium}} & \textbf{Dice}      & 0.784$\pm$0.059  & \textbf{0.786$\pm$0.059}   & 0.773$\pm$0.058    \\
                            & \textbf{Hausdorff} & 32.345$\pm$9.164          & \textbf{31.002$\pm$8.988}  & 33.041$\pm$8.768  \\
                            & \textbf{ASD} & 0.310$\pm$0.171                 & \textbf{0.325$\pm$0.184}   & 0.355$\pm$0.224   \\
\Xhline{2\arrayrulewidth}
\multirow{3}{*}{\textbf{Blood Pool}} & \textbf{Dice}      & 0.909$\pm$0.027 & \textbf{0.908$\pm$0.030}  & 0.901$\pm$0.032   \\
                            & \textbf{Hausdorff} & 38.515$\pm$9.59   & \textbf{38.601$\pm$9.951}        & 39.416$\pm$9.932  \\
                            & \textbf{ASD} & 0.235$\pm$0.200        & \textbf{0.223$\pm$0.201}          & 0.230$\pm$0.204  \\
\Xhline{3\arrayrulewidth}
\multicolumn{2}{c}{\textbf{Compression Rate}}  & 1      & \textbf{$\sim$30x} &  \textbf{$\sim$30x} \\
\multicolumn{2}{c}{\textbf{PSNR (dB)}}              & $\infty$ & $\sim$35  & \textbf{$\sim$36}  \\
\Xhline{3\arrayrulewidth}
\end{tabularx}
\end{table*}

\section{Evaluation}\label{sec:exp}

\subsection{Experiment Setup}
Our proposed machine vision guided 3D image compression framework
was realized by heavily modifying the open-source JPEG-2000 code~\cite{OpenJPEG}. This code also served as our baseline--JPEG-2000 for comparison.  

\textbf{Benchmarks:} we adopted the HVSMR 2016 Challenge dataset~\cite{pace2015interactive} as our evaluation benchmark. This dataset consists of in total 10 3D cardiac MR scans for training and 10 scans for testing. Each image also includes three segmentation labels: myocardium, blood pool, and background. 

\textbf{Evaluation Metrics:} We compared our method with the baseline (JPEG-2000) in following two aspects: 1) segmentation results; 2) compression rate. For the segmentation results, we followed the rule of HVSMR 2016 challenge where the results are ranked based on \textit{Dice coefficient (Dice)}. The other two ancillary measurement metrics, i.e. \textit{average surface distance (ASD)} and \textit{symmetric Hausdorff distance (Hausdorff)}, were also calculated for reference. Among the three metrics, a higher Dice represents higher agreement between the segmentation result and the ground truth, while lower ASD and Hausdorff values indicate higher boundary similarity. 

\textbf{Experiment Methods:} To evaluate our methods comprehensively, two state-of-art segmentation neural network models--DenseVoxNet~\cite{yu2017automatic} and 3D-DSN~\cite{dou20163d} were selected.
We followed the original settings of the two frameworks at training and testing phases but with compressed images. In the testing phase, since the ground truth labels of the selected dataset are not publicly available, we randomly selected five un-compressed training images for training and the rest compressed five for testing. All our experiments were conducted on a workstation which hosts NVIDIA Tesla P100 GPU and deep learning framework Caffe~\cite{jia2014caffe} integrated with MATLAB programming interface.

\subsection{Optimal Parameter Selection}

\begin{figure}[b]
\centering
 \includegraphics[width=0.85\columnwidth]{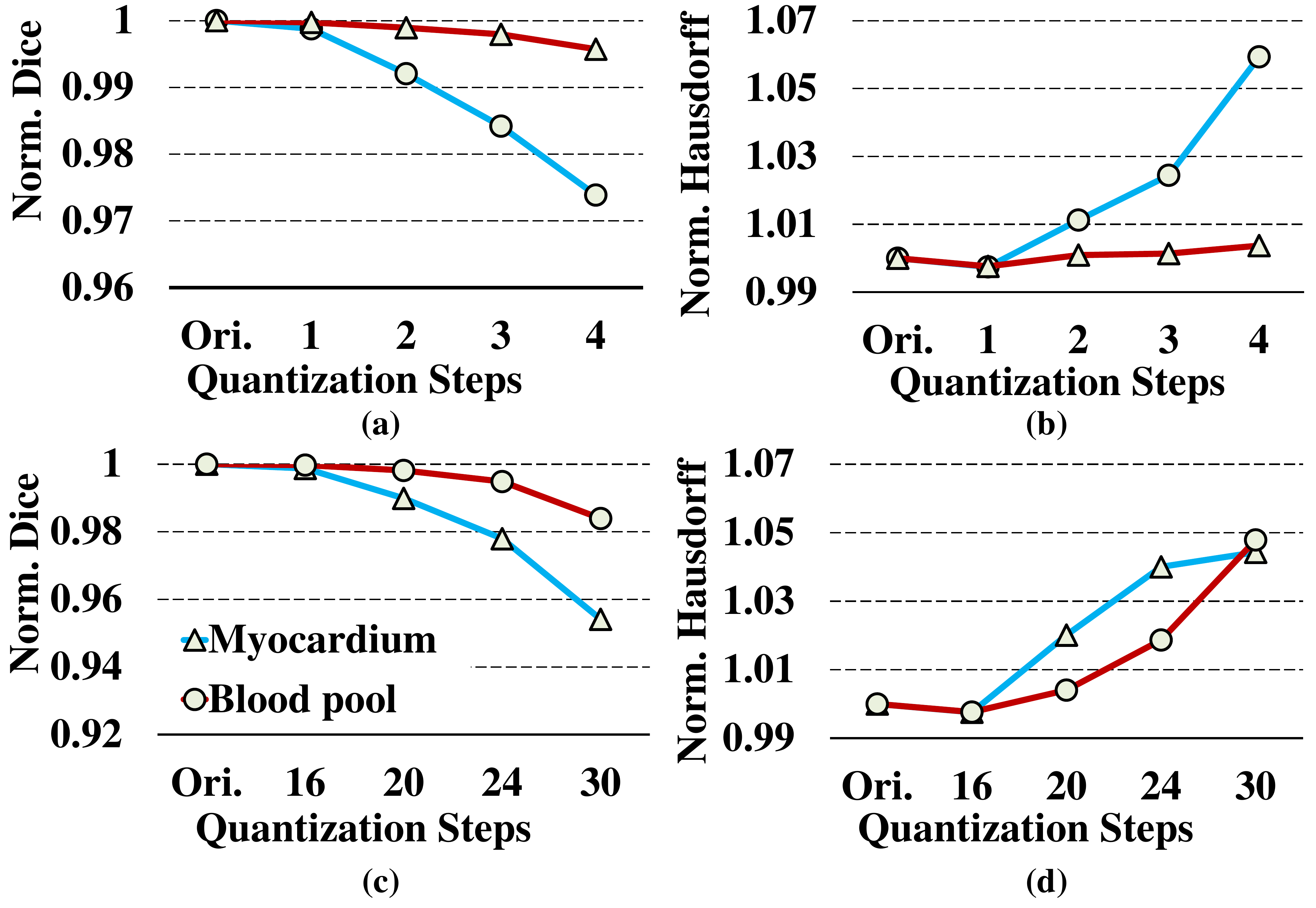}
 \caption{Optimal parameter selection of $Q_{max}$ and $Q_{min}$.}
 \label{opt}
\end{figure}

In this section, we experimentally find the optimal parameters for $Q_{max}$,  $Q_{min}$, $a$ and $b$ in Equation~(\ref{nlm}),  
following the method discussed in Section~\ref{paraOpt}. 

We tested the two cases as discussed in Section~\ref{paraOpt} to find $Q_{max}$, $Q_{min}$. We took normalized dice coefficients and Hausdorff distance as segmentation measurements for an 3D cardiac MR scan and adopted the FCN model--DenseVoxNet. The measurments for two classes-- myocardium and blood pool, are reported. For the lower corner case, as 
Fig.~\ref{opt} (a) and (b) show, the two measurments for both labels do not suffer from any degradation 
only if QS is not larger than 1. Therefore, $Q_{min} = 1$ should be selected as ensure the segmentation results.  
For the upper corner case, the results are shown in Fig.~\ref{opt} (c) and (d).
The two measurements decrease when the QS at $\delta_{min}$ is larger than 16 at both classes, by following a similar trend as the lower corner case.  
Hence, we chose $Q_{max} = 16$ as the upper bound for our quantization step. 
Based on $Q_{max}$, $Q_{min}$ and Equation~(\ref{nlm}), $a$ and $b$ can be decided accordingly.
In our evaluation, we only adopt DenseVoxNet, as an example, to obtain $Q_{max}$, $Q_{min}$ so as to solve $a$ and $b$ in Equation~(\ref{nlm}). 
Then we directly apply it to both DenseVoxNet and 3D-DSN. Note that our method is model agnostic (or rather data specific), since Equation~(\ref{basis}) indicates that the importance of subbands can largely rely on DWT coefficients without correlating with DNN model. Therefore, we can use the same tuned parameters in our compression regardless of network structure. This is also one of the advantages of our method. 

\subsection{Comparison of Segmentation Accuracy}

We first evaluated how our proposed compression framework can improve the segmentation accuracy over the baseline--3D JEPG-2000 using the state-of-the-art segmentation neural network model--DenseVoxNet. For a fair comparison, both our method and 3D JPEG-2000 were implemented at the same compression rate (CR). For illustration purpose, we only report the segmentation accuracy at $CR=30\times$ (results under other compression rates are summarized in the supplemental material). The mean and standard deviation of the three segmentation measurement metrics--Dice, ASD and Hausdorff, are calibrated from the 5 testing images of HVSMR2016 dataset. Note that Dice is the most important metric among the three.

Table~\ref{dens} reports the segmentation results of the two classes--myrocardium and blood pool for the three methods--original (uncompressed, $CR=1\times$), ours and JPEG-2000, under DenseVoxNet. \textbf{First}, the default 3D JPEG-2000 exhibits the worst segmentation results at all the three metrics among the three methods. This is as expected, since JPEG-2000 takes the human perceived image quality as the top priority by offering the highest PSNR ($\sim36$). 
\textbf{Second}, our method, which is developed upon the ``machine vision", can beat JPEG-2000 across all three metrics for both classes, with a lower PSNR ($\sim35$). Impressively, for myocardium, our method can significantly improve Dice, Hausdorff and ASD over JPEG-2000 by 0.018, 2.039, 0.3 on average, respectively. The improvements on blood pool, on the other hand, are relatively limited, given its much higher dice score (0.915 for blood pool v.s. 0.838 for myocardium). 
\textbf{Third}, compared with the original image for both classes, our method only slightly degrades the segmentation results, i.e. $0.001\sim0.004$ on average for Dice, but offers a much higher compression rate ($30\times$ v.s. $1\times$). We also observe that the degradation of all three metrics on compressed images of myocardium (w.r.t. original) is always more significant than blood pool, for both our method and JPEG-2000. This is because myoscardium has a lower dice score than blood pool due to the ambiguous border. These results are consistent with the previous work~\cite{yu2017automatic}.

\textbf{We would like to emphasize that the achieved performance improvement of our method is very significant for segmentation on the HVSMR 2016 Challenge dataset} \cite{wolterink2016dilated}\cite{yu2017automatic} (we also add detailed image by image segmentation results in the supplemental material). Tens of studies performed extensive optimization for segmentation on this dataset. While DenseVoxNet offers the best performance by far \cite{yu2017automatic}, compared with other implementations, it still only improves Dice but degrades Hausdorff and ASD. 
our method, on the other hand, obtains higher performance on all the three metrics on DenseVoxNet.
Furthermore, compared with the second-best method \cite{wolterink2016dilated}\cite{yu2017automatic}, the average improvement of DenseVoxNet on Dice is 1.2\%, while our method can achieve an average improvement of $\sim1.8\%$ for Myocardium on DenseVoxNet.


We also extended the same evaluations to another state-of-the-art FCN--3D-DSN, to explore the response of our method to different FCN architectures. 
As shown in Table~\ref{dsn}, the trend of the results are similar to that of DenseVoxNet, except for lower segmentation accuracy. Note this is caused by the neural network structure difference, and DenseVoxNet currently achieves the state-of-the-art segmentation performance. As expected, again, our method significantly outperforms JPEG-2000 at the same compression rate ($30\times$) across all the three metrics, i.e. 0.013 (myoscardium) and 0.007 (blood pool) on average for dice score, while providing almost the same segmentation performance as that of uncompressed version--original ($1\times$). These results clearly show the generalization of our method. 




It is also notable that from both tables, the segmentation results from compressed images using our method sometimes even outperform that of original images. This is because compression as frequency-domain filtering also has denoising property. Although the training process attempts to learn comprehensive features, the importance of the same frequency feature may vary from one image to another for a trained DNN. As a result, after compression, the segmentation accuracy of some images may be improved because the unnecessary features that can mislead the segmentation are filtered, as demonstrated in Fig.~\ref{motivationFig}(b) (Our method is better than Original CT). For most images, the segmentation accuracy after compression is still slightly degraded compared with the original images due to minor information loss at high compression rates, though our compression method tries to minimize the loss of important features.

\subsection{Comparison of Compression Rate}
\begin{figure}[b]
\centering
 \includegraphics[width=0.95\columnwidth]{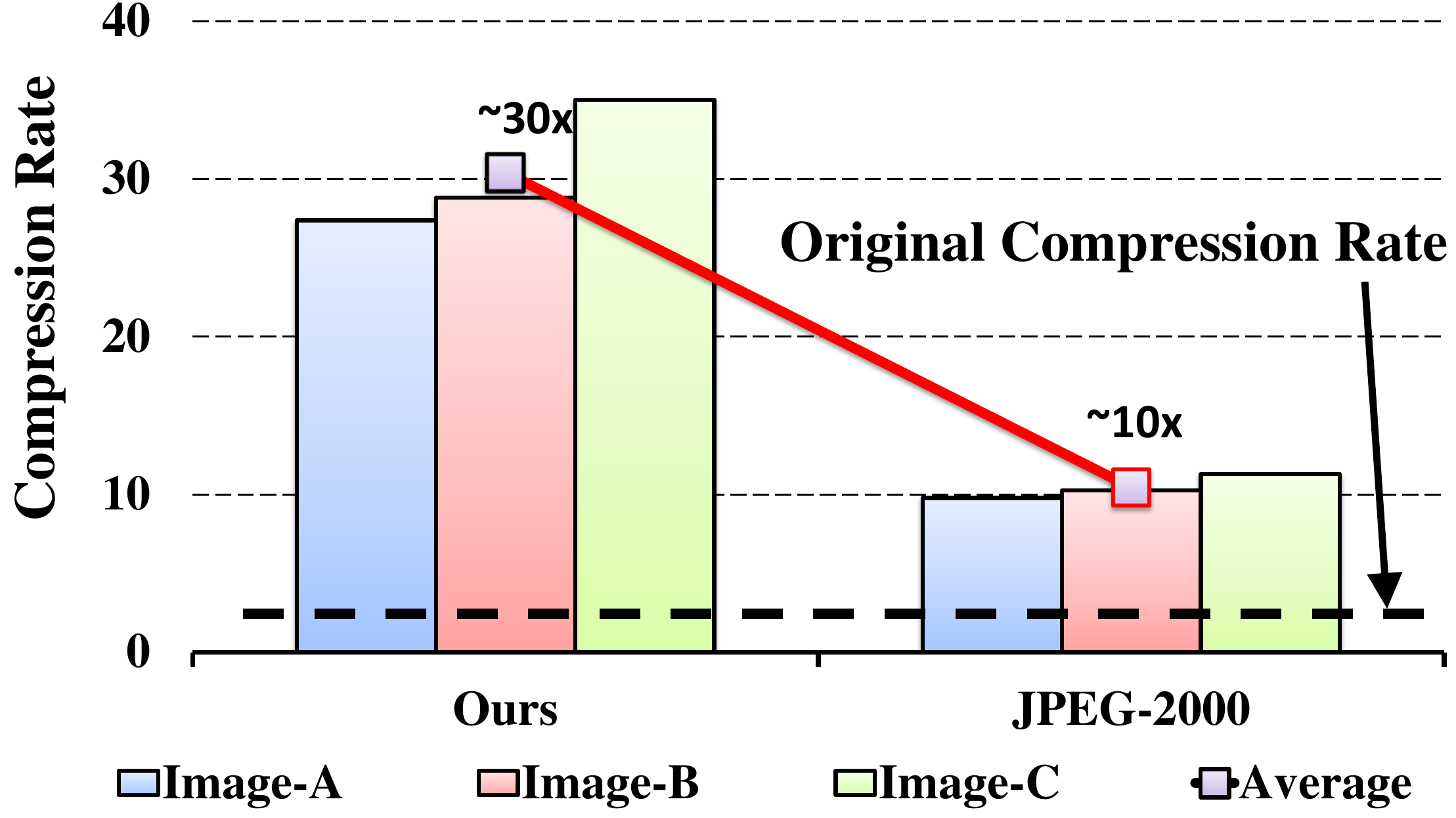}
 \caption{Compression rate comparison of our method v.s. JPEG-2000 under the same segmentation accuracy.}
 \label{comp}
\end{figure}

In this section, we explore to what extent our proposed machine vision-oriented compression framework can improve the compression with regard to the human-visual based 3D JPEG-2000, for medical image segmentation.  For a fair compression, we compared the compression rate (CR) of these two methods under the same segmentation accuracy for myocardium using DenseVoxNet. Dice score (0.834) was selected as it is the prime metric to measure the quality of image segmentation. Since the compression rate may vary from one image to another, we chose three representative images from the dataset. As Fig.~\ref{comp} shows, our method can always deliver the highest compression rate across all the images. On average, it achieves $30\times$ compression rate over the original uncompressed image. Compared with 3D-JPEG 2000, 
our method can still achieve $3\times$ higher image size reduction, without degrading the segmentation quality. Still taking the example from section~\ref{int}, we assume the transmission time of a 3D CT image of size 300MB via fixed broadband internet ($22.79Mb$) to cloud is 13s, while the image segmentation computation time on cloud is merely 100ms. Putting these two together, a single image segmentation service time on cloud for our method ($30\times$) and JPEG-2000 ($10\times$), are 0.53s and 1.4s, respectively, translating into $2.6\times$ speed up.

\subsection{Overhead}
Our method is built upon 3D-JPEG 2000 by only adding two simple operations: standard deviation calculation for 16 subbands
and 
equation set solution (Equation (\ref{nlm})) with only four variables.
Since we reuse the majority of JPEG-2000's function units, the compression and decompression time are at the same level as that of JPEG-2000, e.g., 0.12ms for a 512$\times$512 image \cite{matela2009gpu}, which is almost negligible compared with image transmission and segmentation time.    
Therefore, 
we expect that our light-weighted machine vision guided 3D image compression framework can find broad applications in medical image analysis.

%% file: 6.conclusion.tex
\section{Conclusion}
Due to the high computation complexity of DNNs and the increasingly large volume of medical images, cloud based medical image segmentation has become popular recently.
Medical image transmission from local to clouds is the bottleneck for such a service, as it is much more time-consuming than neural network processing on clouds. 
Although there exist a lot of 3D image compression methods to reduce the size of medical image being transmitted to cloud hence the transmission latency, almost all of them are based on human vision which is not optimized for neural network, or rather, machine vision.
In this paper, we first present our observation that machine vision is different from human vision. Then we develop a low cost machine vision guided 3D image compression framework dedicated to DNN-based image segmentation by taking advantage of such differences between human vision and DNN. 
Extensive experiments on widely adopted segmentation DNNs with HVSMR 2016 challenge dataset show that our method significantly beats existing 3D JPEG-2000 in terms of segmentation accuracy and compression rate.

